\documentclass[conference]{IEEEtran}

\ifCLASSOPTIONcompsoc
  \usepackage[nocompress]{cite}
\else
  \usepackage{cite}
\fi

\ifCLASSINFOpdf
\usepackage[pdftex]{graphicx}
\else
\fi

\usepackage[float]{}

\usepackage{algorithm}
\usepackage{algpseudocode}
\usepackage{textcomp}
\usepackage[english]{babel}
\usepackage[utf8x]{inputenc}
\usepackage{amsmath}
\usepackage{amssymb}
\usepackage{graphicx}
\usepackage[export]{adjustbox}
\usepackage{array,multirow}
\usepackage[table]{xcolor}
\usepackage{listings}
\usepackage{cleveref}
\usepackage{url}
\usepackage{todonotes}
\graphicspath{ {figures/} }
\makeatletter
\DeclareTextCommandDefault{\textleftarrow}{\mbox{$\m@th\leftarrow$}}
\makeatother

\begin{document}

\title{Project Thyia: A Forever Gameplayer}

\IEEEoverridecommandlockouts
\IEEEpubid{\begin{minipage}{\textwidth}\ \\[12pt]
978-1-7281-1884-0/19/\$31.00 \copyright 2019 IEEE
\end{minipage}}

\author{
\IEEEauthorblockN{Raluca D. Gaina, Simon M. Lucas, Diego Perez-Liebana}
\IEEEauthorblockA{\textit{Game AI Research Group}\\
Queen Mary University of London, UK\\
\{r.d.gaina, simon.lucas, diego.perez\}@qmul.ac.uk}
}

\IEEEtitleabstractindextext{%
\begin{abstract}
The space of Artificial Intelligence entities is dominated by conversational bots. Some of them fit in our pockets and we take them everywhere we go, or allow them to be a part of human homes. Siri, Alexa, they are recognised as present in our world. But a lot of games research is restricted to existing in the separate realm of software. We enter different worlds when playing games, but those worlds cease to exist once we quit. Similarly, AI game-players are run once on a game (or maybe for longer periods of time, in the case of learning algorithms which need some, still limited, period for training), and they cease to exist once the game ends. But what if they didn't? What if there existed artificial game-players that continuously played games, learned from their experiences and kept getting better? What if they interacted with the real world and us, humans: live-streaming games, chatting with viewers, accepting suggestions for strategies or games to play, forming opinions on popular game titles? In this paper, we introduce the vision behind a new project called Thyia, which focuses around creating a present, continuous, `always-on', interactive game-player.
\end{abstract}
}

\maketitle

\IEEEdisplaynontitleabstractindextext

\section{Introduction} \label{sec:intro}

The quest for artificial general intelligence (AGI) has been pursued for many years. Yet ``no free lunch" stands true to this day~\cite{ashlock2017general} and there exists no one method that is able to solve all problems. Some researchers have been trying to model the human brain in order to give algorithms the power of learning that humans have~\cite{banino2018vector}. Generally, humans are fairly good at learning how to perform a variety of tasks ranging in difficulty, from using our body (walking, picking things up, dancing), to maths (counting, fractions, solving equations), to producing creative works (writing, drawing, painting, designing games) and all in-between. Not all of us are the best at all the different tasks, but most of us are fairly good at various tasks within the same domain and at figuring out how new problems work given the knowledge base built over our lifetime.

Although a lot of important advances in AI have been made in games, and games are still actively used as testing environments for AI, algorithms are only able to solve (in this case, win or achieve a high score) a subset of existing games~\cite{perez2018general}. Planning and learning algorithms alike are unable to act in an intelligent manner in all given games, unless they use human-tailored heuristics or features (often game-specific). They do excel in some games, and different methods are better at different types of tasks. 
Here, we look at it from the perspective of human intelligence: humans don't only learn, or don't only plan, when faced with a new problem. They plan based on existing knowledge, execute the plan, and use the new experiences to update their knowledge. We believe that combining planning and learning methods is key to AGI.

We do notice one drawback in game-playing AI research that is rarely addressed, to our knowledge. The usual steps of running a game-playing AI algorithm are as follows:
\begin{enumerate}
    \item Write / obtain algorithm.
    \item Set up problem domain.
    \item Press run.
    \item Run ends with some result, maybe statistics.
    \item Instance of AI no longer exists. \label{list:AIend}
    \item Rerun new instance of AI for new result.
\end{enumerate}
Point~\ref{list:AIend} here is where our interest lies. Even in the case of learning algorithms, they run for a limited number of steps or episodes, however long the researcher can afford to spend testing the method (in either time or money). The algorithm may converge in the given time, therefore, even if given longer, it wouldn't do any better, but often it does not. If any bugs are found or thought to exist, the knowledge acquired previously is scrapped and it all starts from zero again. 
As most algorithms are stochastic, it's possible to store the random seed of a good run to reuse later on; we consider this to be a small attempt at copying the AI instance, instead of preserving it. 

A different scenario is presented by bots hosted on servers and interacting with humans, such as general chat bots, Twitter or Slack bots~\cite{veale2018twitterbots}. These bots are given mostly social intelligence, so that they can respond to human requests, or maybe even initiate conversation (more rarely). We take from these the concept of (almost) permanent existence in the cyberspace, as well as their ability to interact with humans. 

A similar example from the game AI domain is ANGELINA, the game designer~\cite{cook2018redesigning}. Unless Michael Cook decides to take her down for updates or a break, she continuously and autonomously creates games, improves them, names them, or throws them away if she decides they're unworthy. She also sometimes interacts with humans, either through Twitter messages, streaming the design and test process on Twitch or by simply sharing the games it creates\footnote{\url{https://gamesbyangelina.itch.io}}. In~\cite{cook2018redesigning}, Cook and Colton discuss the benefits of continuous creative systems, highlighting long-term growth and development.

In this paper, we present Eileithyia (or Thyia for short), the game player. Inspired not only by previous research and internet trends, but also by Greek mythology. Eileithyia is the Greek Goddess of childbirth, or life, as this would be a more interesting framing in this context. 
Thus Thyia is now Goddess of artificial life in a game-playing context, a system combining learning and planning. Thyia would act in a similar way to ANGELINA: she exists in cyberspace (or, simply put, continuously running on a server) and her purpose in her potentially endless life is to play games and become the best player humans have seen. Thyia uses planning methods to play games, informed by the knowledge she gathers over her lifetime, and learns from her experiences to improve her performance over time.

We highlight that Thyia would be the first step towards combining multiple areas of research and increasing their presence and potential impact in the real world. We envision that game-playing agents would be able to use the knowledge and experience gathered by Thyia to further improve their own performances even in time-limited scenarios. Thyia is meant to showcase the true strength of modern techniques when used together for long-term development.

We can summarise our contributions as follows. We propose a new way of thinking about game-playing Artificial Intelligence as entities that exist in cyberspace. We introduce Project Thyia, which centres around such an AI entity. We combine existing planning and learning methods to allow Thyia to not only plan through games, but also learn from experiences and improve over time, by using knowledge acquired as well as by tuning its large parameter space and structure. Finally, we discuss difficulties imposed by a continuous game-player.

The paper will first review related literature in Sections~\ref{sec:lit1}, \ref{sec:lit2}, \ref{sec:lit3} and \ref{sec:lit4}. The concepts behind Project Thyia are described in Section~\ref{sec:thyia} and Section~\ref{sec:ethics} addresses ethical concerns related to the project. We conclude in Section~\ref{sec:end} with takeaways and more possibilities for expansion of the multi-disciplinary project proposed.


\section{AI entities} \label{sec:lit1}

Creating artificial life that we, as humans, can interact with in meaningful ways is the topic of many books and films, with ongoing research trying to make such concepts into the real world. We will explore some of the advances in this area in this section, with a focus on interactive `always-on' AI. 

Perhaps a most commonly accessible form of interactive AI is chatter bots. They are generally AI algorithms running continuously on a server, accepting some form of human input (i.e. text or speech) and returning some output in response to the input received. Conversational bots~\cite{schlesinger2018let} are largely based on natural language processing and databases of appropriate responses which could be manually designed or automatically extracted~\cite{huang2007extracting}. An early program with such intent is Weizenbaum's Eliza~\cite{weizenbaum1966eliza}, a chatterbot built to respond to certain keywords in order to facilitate communication between man and machine. This concept is also used in the development of platform-specific bots, such as Twitter bots, which may post various content on its intended platform with the aim of interacting with other users~\cite{veale2018twitterbots}.

Some games adopt this concept and research on social, interactive characters in order to create more immersive experiences for their players. A great example here is Emily Short's game ``Galatea''~\cite{short2000galatea}, which focuses on interactive storytelling. Emily Short writes about NPC conversation systems~\cite{short2011if}, showcasing different functionalities these can take. We are interested in the tutorial system most, although in our case it would be reversed: the humans would be giving the AI hints, and not the other way around, as in Matt Wigdahl’s ``Aotearoa''~\cite{wigdahl2010aotearoa} or Santiago Onta\~non's ``SHRDLU''~\cite{ontanon2018shrdlu}. These characters, although becoming more and more impressive with the inclusion of memory, personality and adaptation to different players and play styles, they do exist only within the game. Our vision wishes to take this concept further and bring more presence into the real world to such characters.

Darius Kazemi, known as Tiny Subversions\footnote{\url{http://tinysubversions.com}}, is an internet artist who creates interesting `continuous' bots. One example is his ``Random Shopper'' project, which consists of an AI entity that buys a new random item from Amazon every month, within a certain budget, and has it shipped to Kazemi's home for a regular surprise. 

Zukowski and Carr recently used Deep Learning to create an AI entity which live-streams music it generates on YouTube~\cite{zukowski2018generating}. The virtual band, Dadabots, creates death metal pieces with the aims of showing that AI is able to capture interesting differences between various music genres.

Perhaps the most notable AI entity with a presence in the real world from the games domain is ANGELINA. Initially created as an automatic system for designing entire games, with a focus of exploring the limits of a software's creativity and novelty while creating interesting playable experiences~\cite{cook2017angelina1,cook2017angelina2}. Cook and Colton describe in~\cite{cook2018redesigning} the extension of their vision for autonomous continuous game creation, with a highlight to the opportunities this methodology opens for long-term improvement. We base our concepts largely on their ideas, extending further to a multi-faceted game-playing system.


\section{Continual learning} \label{sec:lit2}

An interesting research area which addresses similar problems to our domain is that of continual learning (also known as lifelong learning, or sequential learning), which focuses on long-term learning for continuous development, often on a sequence of different tasks. One definition of this domain is the study of agents capable of interacting with their environment (in our case, a game), with limited computational resources, that is started once and run for a long time (once started, no more changes are allowed) with the aim of continuously improving at fulfilling its goals over a period of time~\cite{parisi2019continual}. The main differences to our system are the lack of changes allowed once it starts (we wish to allow for updates and changes in the modules part of our system), as well as the lack of interaction with the wider world outside of the agent's environment (we wish our agent to not only be getting better at the games it plays, but also to have a presence in the real world).

However, a lot of the concepts described in continual learning research can be applied in our case as well. This section will review several recent works with relevant and interesting results and/or takeaways.


Parisi et al.~\cite{parisi2019continual} review several works in the area. They note that current approaches are still facing several issues, including flexibility, robustness and scalability. Interestingly, most learning models rely on large amounts of annotated data to function in supervised domains. We wish to emphasise our focus on efficient learning and gathering of data for learning from our planning agent, as well as the modularity of our proposed system which would allow for new games to be added in, which may not respect the same assumptions of our current corpus of games. Thus flexibility and robustness are key aspects we consider, with scalability to more complex game domains an interesting path for future developments.



One example of applying continual learning methods to the games domain is the work by Schwarz et al.~\cite{schwarz2018progress}, who propose an algorithm which compresses its memory after learning each new task so as to preserve key concepts, both old and new. They test their method on the Atari suite and show it to be better than other knowledge preservation methods like Elastic Weight Consolidation (EWC)~\cite{kirkpatrick2017overcoming} on several games. These methods, as well as those in \cite{lopez2017gradient}, \cite{chaudhry2018efficient} and others can be used to enhance our learning component, although in this paper we choose to focus on simpler Neural Network approaches.

As highlighted by Diaz et al.~\cite{diaz2018don}, most of the focus on continual learning is on memory retention, shaping the knowledge acquired and selectively deciding what, when and how to expand the knowledge, so as to improve performance on new tasks without affecting previously learned ones. Diaz et al. suggest that evaluation of these methods is also very important and propose using several metrics applied at intervals during the learning process: accuracy, knowledge transfer, memory and computational efficiency. They combine these into an overall weighted-sum score based on which they can rank various methods on the iCIFAR-100 dataset. 
We consider these notes important for future evaluations of our system.


\section{Learning while planning} \label{sec:lit3}

There has been increasing interest in the research community regarding the combination of learning and planning methods. Each have their own strengths and weaknesses. Learning methods not only require some form of game state feature extraction in order to be able to play games; but they also need significant resources for training and they lack generalisation across different tasks~\cite{mnih2013playing}. Planning methods have proven to be very good at a variety of different tasks and they can work online, with no training, and in real time; but they do require a model of the game in order to simulate possible future scenarios, and they struggle in sparse reward environments~\cite{gaina2019sparse-rewards}. One way forward is to draw upon the strengths of both techniques in order to build a fast, effective and general algorithm.

There have been several advances in this direction in board games, where the approach is to use a search algorithm (often Monte Carlo Tree Search) as an expert to generate gameplay data, and a learning algorithm (often a Deep Neural Network) which uses the gameplay data to train and perform better than either algorithm would individually. A prime example is AlphaGo~\cite{Silver2016}, followed by AlphaGoZero~\cite{silver2017mastering} and AlphaZero~\cite{silver2018general}, all of which combine Monte Carlo Tree Search (which generates gameplay data) and Neural Networks (which use the gameplay data to train policies and value estimates) to successfully beat the state of the art in the game Go (and Chess and Shogi in the case of the latest AlphaZero program).

Anthony et al.~\cite{anthony2017thinking} split the task of playing the game of Hex into two areas: planning efficiently and generalising the plans across different boards and opponents. They use tree search to plan, aided by a neural network policy to guides the search, and Deep Learning to further generalise the plans. They pinpoint the benefits of their approach and the great results of combining the two approaches, which mean the agent is capable of winning against previous champions.

These ideas were further developed and applied to video games, which differ mainly through their real-time aspect, as well as increased complexities of dynamic worlds: AI methods only have a limited time to make decisions. In this paper, we also focus on video games, although the concepts described could be extended to any games or problems. Jiang et al.~\cite{jiang2018feedback} apply a combination of Monte Carlo Tree Search (MCTS) and Neural Networks to obtain a competitive ``King of Glory'' player, where the heuristic used by MCTS to evaluate leaf nodes is improved incrementally based on the results returned. 

Lowrey et al.~\cite{lowrey2018plan} developed a framework based around the idea of combining planning and learning, called POLO, which consists of several continuous control tasks, such as humanoid locomotion or hand manipulation. Most interestingly, they suggest that this combination of methods brings several benefits including reducing the planning horizon, while finding good solutions beyond the local space currently being explored.

A particularly relevant recent work is a combination of Neural Networks (NN) and Rolling Horizon Evolutionary Algorithms (RHEA) applied to a series of MuJoCo control tasks~\cite{tong2019rhea}. Tong et al. apply the idea used in AlphaGo works by replacing the MCTS with RHEA, and using NNs to generate policies and state value estimations. The policies are used to initialise RHEA, while the state value estimations are used in the fitness evaluations of plans generated by the evolutionary algorithm. The network is then updated after the evolutionary process completes in order to improve value estimations and, implicitly, the policies as well, so that the whole system learns to generate better plans over time. They suggest their method is efficient in learning interesting behaviours. We adopt these ideas into the learning part of our system with several extensions, as detailed in the next section.


\section{Planning to Learn} \label{sec:lit4}

With a different perspective, we consider active learning as a form of increasing the autonomy of our system. Within pattern classification and language learning, active learning is used to describe cases where the learning agent generates its own patterns to submit to an Oracle which then informs the learner of the class of pattern~\cite{settles2009active}. Active learning algorithms can use this approach to formulate highly relevant queries that may enable more sample efficient learning.  

Reinforcement learning (RL) agents are already in control of their own destiny within a game, since what they experience depends on the actions they take. In particular, RL algorithms can use concepts such as intrinsic motivation (which may be related to novelty search) to take an active approach to learning in the absence of sufficient external reward signals. This has been used to good effect to boost performance on games with sparse reward landscapes such as Montezuma's revenge~\cite{ecoffet2019go}.

Beyond this, we envisage an even more active type of RL in at least two cases: setting in-game scenarios in order to improve performance on a particular game, and deciding which games to play in order to maximise ``personal'' development. In the first case, and agent would reason about its current state of ignorance, and set up scenarios to test various hypotheses. Any game that allows user-defined levels would support a form of this, though the ideal would be to have a set of learning games with an active interference API (Application Programming Interface), allowing for alterations of the game state mid-game in order to explore specific consequences. This would enable a much more direct manipulation of the game-state than could be achieved by the normal process of taking actions within a game, in turn leading to more sample-efficient learning of highly performant strategies. 

In the second case, the agent would analyse its more general short-comings and select games with which to hone its skills. As far as we are aware, neither approach has been tried within the field of AI and Games.



\section{Proposed System: Thyia} \label{sec:thyia}

\begin{figure} [ht]
    \centering
    \includegraphics[width=\columnwidth]{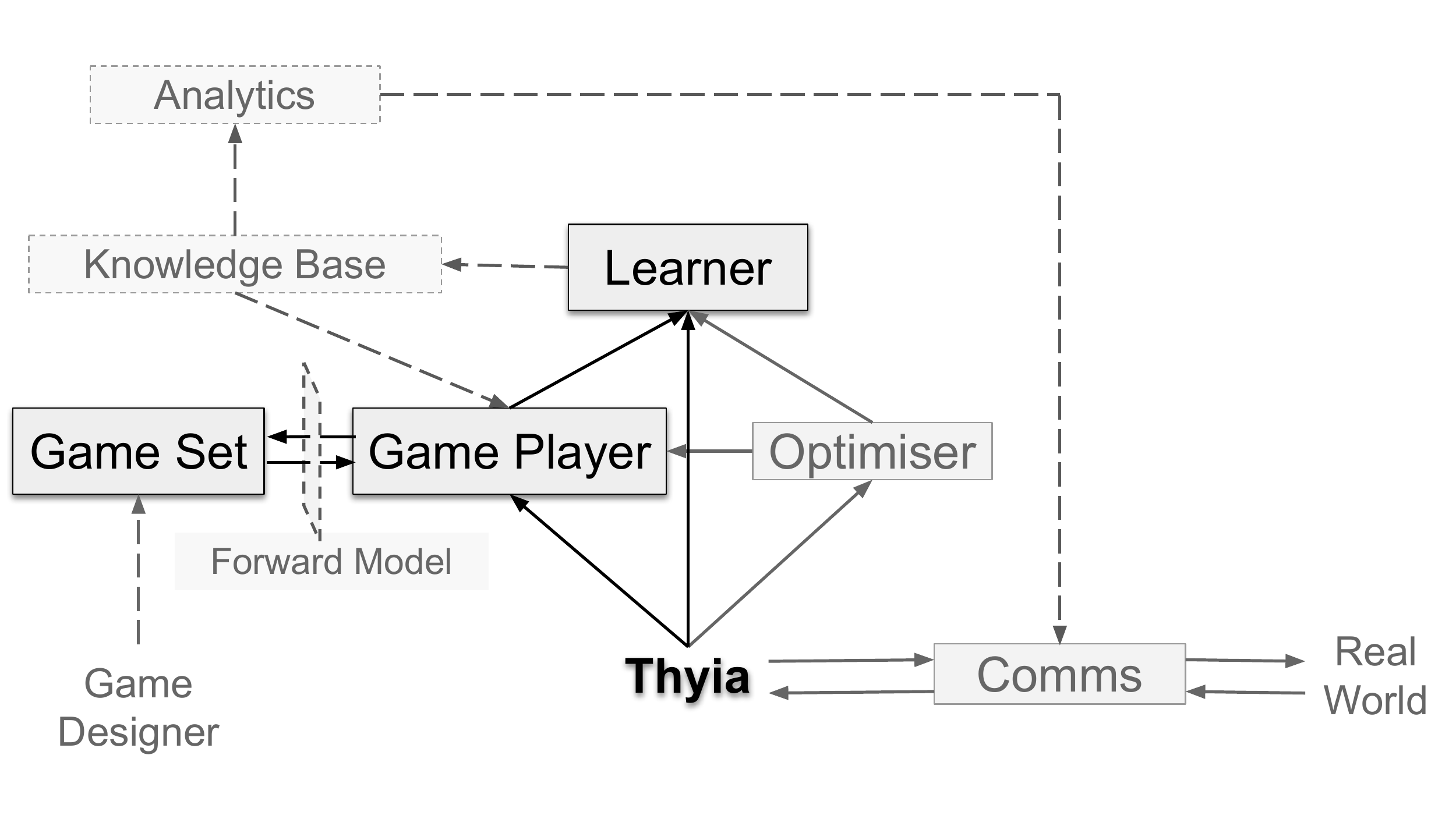}
    \caption{Thyia system. Composed of 3 core modules: a game player (planning AI agent), a learner and a game set. Additional modules include external communications with the ``real world'' for game sharing and human interaction; and an optimisation module for tuning the game player and learner's parameters. Further enhancements include a knowledge base, detailed analytics and forward model learning. We include a possible connection with a game designer, which would be providing games for the system to play.}
    \label{fig:thyia}
\end{figure}

Thyia is a large system comprised of several modules. The modularity aims to allow for different parts to be maintained, improved, updated or rebooted independently, in order to avoid potential loss of data in the larger system. See Figure~\ref{fig:thyia} for an abstract representation of the system envisioned in this paper.

\subsection{Game Set Module}

As we are targeting a planning agent which also learns over multiple runs, an essential part of the system is the set of games. Multiple frameworks for general-purpose game-playing exist. One of the earliest general game frameworks was Metagamer, which used a Game Description Language to define the rules of chess-like games (i.e. Chess, Chinese Chess, Checkers, Draughts and Shogi) and automatically generate variations for game-players to attempt to beat~\cite{pell1993strategic}. A similar project was developed by Jeff Mallett and Mark Lefler, called Zillions of Games\footnote{\url{http://www.zillions-of-games.com/ZOG.html}}. They expanded the range of games to general board games, using a LISP-like language for game definition. The AI received a lot of information about the game: not only the actions available, but also the board structure and the goals of the game. Humans could use the system to not only create new games, but they could also choose to play their games against an AI using alpha-beta pruning and transposition tables.

This idea evolved further into the General Game Playing competition (GGP)~\cite{genesereth2005general}, which includes turn-based deterministic board and puzzle games and offers the entire rule set to the agents in order for them determine their strategy. The Arcade Learning Environment (ALE)~\cite{bellemare2013arcade} improves on this challenge with their focus on real-time deterministic arcade games. ALE in particular has received much attention in the last few years from AI researchers~\cite{guo2014deep,machado2018revisiting,ecoffet2019go}. 

However, we choose to use the General Video Game AI framework (GVGAI)~\cite{perez2014gvgai}, which comprises of a large (and increasing further on a yearly basis) set of games with different properties of interest, such as partial observability, stochasticity and a variety of game mechanics and score systems (including dense and sparse rewards). The method proposed in this paper could be used in any of the GVGAI games, as they all use a common interface and no game-specific knowledge is embedded in our system - thus it presents an emergent quality for generalisation across games. Additionally, GVGAI uses the Video Game Description Language (VGDL) to define its games, which allows for easy creation of new games or variations of existing ones, leading to a potentially infinite supply of games varying in features and complexity.

A possible line of work to increase the presence of this system would be an integration with a game designer (e.g. ANGELINA~\cite{cook2018redesigning}). It follows naturally that ANGELINA and Thyia could make an excellent team, one creating games based on the results and feedback of the other which plays them.

\subsection{Planning Module}

The core part of the system will be the actual game-player: the algorithm which is able to play unknown games. We choose to base the game-player on a planning method, due to their flexibility, adaptibility and lack of training necessary, as well as high performance across multiple games~\cite{perez2018general}. The downsides of these methods are two-fold: they do not usually learn between games (i.e. the performance of the method is likely to be the same the first and the hundredth time it plays a game, save for game or algorithm stochasticity) and they require a game model to be able to simulate possible future states. The first issue is addressed throughout this paper; the second will be discussed further in Section~\ref{sec:end}.

Although Monte Carlo Tree Search is a commonly used in game-playing research~\cite{Browne2012}, as well as some commercial games (e.g. Creative Assembly's ``Total War: Rome II''~\cite{champandard2014monte}), recent work has shown Rolling Horizon Evolutionary Algorithms (RHEA) to achieve a higher performance in many games~\cite{gaina2017enhancements}, as well as offering many customisation opportunities: the underlying Evolutionary Algorithm can ranged from a 1+1EA~\cite{lucas2018n}, compact GA~\cite{lucas2017efficient}, CMA-ES~\cite{tong2019rhea} and population-based GA~\cite{gaina2017analysis}. We choose to start with RHEA for the planning module, but other methods such as MCTS could also be easily integrated.

Many RHEA hybrids can be created for interesting and diverse results~\cite{gaina2017population,gaina2017enhancements}, while the idea behind the algorithm remains the same: evolving sequences of actions at every game tick and evaluating each sequence with the use of a game model, to simulate through the actions and assess the final state reached by following the given actions. The value of this state, given by a heuristic, becomes the fitness of the individual. At the end of the evolutionary process (when some budget has been reached), the first action of the best individual is chosen to be played in the game.

In our implementation, RHEA includes over 30 parameters, allowing for not only operational settings to be modified (i.e. individual length, mutation rate), but also the very structure of the algorithm (keeping the population evolved from one game tick to the next with a shift buffer, including or excluding evolutionary operators, adding Monte Carlo rollouts at the end of the individual when evaluating, etc.). These options are all collected from past literature~\cite{gaina2017analysis,gaina2017population,gaina2017enhancements,gaina2019sparse-rewards} for a resulting EA with a parameter search space size of $1.741E12$.

The internal state of the planning agent can be fully represented by the random seed and its parameter settings (i.e. given parameters and random seed, the exact same behaviour would be achieved in multiple runs of the algorithm). Therefore, in order to preserve the internal state of the planning module, we need to store its parameter settings and random seed. In the case of updates being required, we would then be able to pause the system, disconnect the planning module, perform any updates and hook it back in with access to its previous parameters and seed. In order to avoid complications with changes in parameter space, the parameter space itself is built separately and modularily, so as new parameters may be added in without impacting any other part of the system.

\subsection{Learning Module}

Given the success of several works of combining planning methods with Neural Networks (NN), the learning part of our system would also take the form of a NN. However, there are two main aspects to consider here: the state representation and the network architecture. In GVGAI, the typical state representations used are: grid observations (NxM matrix with each cell representing one or multiple sprites at that location)~\cite{braylan2016object,kunanusont2017general,Narasimhan2017} or compressed feature vectors extracted automatically from an image representation of the state with an Object Embedding Network~\cite{Woof2018}. Various learning methods are used with different architectures, such as: Value Iteration Networks~\cite{Narasimhan2017}, Q-Learning~\cite{Woof2018} and Deep Reinforcement Learning~\cite{kunanusont2017general,Torrado2018}. We propose using an architecture similar to that of the successful AlphaGo~\cite{silver2017mastering}, although the increased game complexity should be taken into account.

Due to the aims of a general game playing, a limitation to consider is differences in games which make it harder (if not impossible) to apply a model learned in one game to another. In the simplest case, the agent would be told whether it is playing a different game, and learn different models per game. However, in order for the system to be autonomous, not relying on human information, as well as learn efficiently, another key aspect to consider for the learning module is transfer learning or extraction of key concepts which are generally applicable across games (i.e. walking into walls is generally not allowed) similar to the work of Narasimhan et al. in~\cite{Narasimhan2017}, for example.

To describe the internal state of the NN, we would need to save the generated model in order to be able to rerun the exact same instance of the algorithm.

\textbf{Combining planning and learning.} There are various ways in which we can combine the planning and learning approaches. Our proposed method extends from the AlphaGoZero~\cite{silver2018general} and p-RHEA~\cite{tong2019rhea} approaches in literature, as detailed below.

\begin{itemize}
    \item \textbf{Initialisation}: We replace the uniform distribution used in the random population initialisation with external distributions provided by the NN. Starting from the current game state $S_t$, we query the NN for the action distribution $\pi_t$. This policy is followed using the softmax function and the next state is simulated according to the action $a_t$ selected from $\pi_t$, giving us $S_{t+1}$. The same process is repeated until we generate a full action sequence of the desired length for use within RHEA. We then generate the rest of the individuals in the initial population as mutations of the first.
    \item \textbf{Mutation}: We replace the uniform distribution used in selecting a new value for a gene $g$ being mutated with an external distribution provided by the NN. Given the game state obtained by simulating through the sequence in the individual gene $g$ belongs to, up until (and including) gene $g$, noted as $S_g$, the NN return the action distribution $\pi_g$. We modify $\pi_g$ to set the weight of the current value of gene $g$ to 0, and we perform weighted sampling from this distribution to obtain the new value for gene $g$ (guaranteed different from previous).
    \item \textbf{Fitness}: We include in the individual fitness the external state value $v_t$ provided by the NN, weighted by $\alpha$ (see Equation~\ref{eq:fitness}, where $R_{value}$ is the rollout value obtained by RHEA individual evaluation and $N_{value}$ is the NN value output for the final state reached through the rollout).
\end{itemize}

\begin{equation}
    f = (1 - \alpha) \times R_{value} + \alpha \times N_{value}
\label{eq:fitness}\end{equation}

\subsection{Optimisation Module}

A different line of work in improving the performance of game-players (as opposed to relying on learning) is the optimisation of their parameters. Given the large parameter search spaces for both the RHEA and NN components of our system, it is highly unlikely that a human user would be able to select the perfect combination of parameters which would result in the highest performance, or most efficient learning. Therefore, we add an optimisation module to Thyia. Although several optimisation methods have been explored in literature and any could be integrated with our system, we choose to focus on the N-Tuple Bandit Evolutionary Algorithm (NTBEA)~\cite{kunanusont2017n}, which has been shown to perform well and robustly on various problems, in tuning game parameters~\cite{kunanusont2018scoretrend} as well as game-player parameters~\cite{lucas2018n,lucas2019efficient,bravi2019rinascimento}.

NTBEA is a model-based optimiser based on an Evolutionary Algorithm, which uses bandit-based sampling and detailed statistics on combinations of parameters in order to optimise hyper-parameters. It highlights fast convergence in noisy optimisation problems even with small computational budgets and it scales well for large search spaces~\cite{lucas2019efficient}, although it has yet to be tested on a search space as large as Thyia's. 


The addition of this module does raise additional difficulties for the learning system: since the parameters of either the playing or learning algorithm would change, the data received by the learner could vary significantly in terms of the game-player's behaviour. Therefore, the learner needs to be general enough to not make any assumptions in the data it receives, in order to be able to cope with high variations.

\subsection{Human interaction module}\label{subsec:interaction}

Another important part of the vision for this system is the ability to interact with the ``real world''. Ultimately, the goal of AGI is to bring benefits to humans in a multitude of real-world problems, thus the humans must be brought into the loop. There are several ways in which this could be achieved.

\textbf{Direct interaction.} One of the clearest cases would be a direct communication between the user and the AI, where the AI would use natural language processing to understand humans and reply to them intelligently (or, in the simplest case, building a database of keywords which trigger certain responses from the system). The purposes of this interaction form could vary from asking Thyia to play certain games, or asking for its thoughts on games its played. This leads to further extensions of AI able to form opinions supported by solid arguments or facts. Additionally, given its support for multi-player games, humans could play alongside the AI for another form of direct interaction.

\textbf{Knowledge and statistics display.} Depending on the specific chosen representation for Thyia's knowledge, it may be hard to understand by humans: it could end up being an endless string of numbers which would mean little to us without the capabilities of fast computation. Therefore, visualising statistics about the games, the game-player's behaviour or the knowledge gathered would be interesting and useful to gain a better understanding about the system's inner-workings.

\textbf{Live streaming.} A common way for human game-players to interact with others is through live streams (e.g. via popular platforms like Twitch and YouTube) and video sharing. There is a large community which revolves around the concept of sharing gameplay with others that watch and comment on the game being played, suggest strategies for the game or offer helpful information. In a similar way, Thyia could be streaming the games it plays to enter this community of human game-players, while opening a direct communication channel through which it could even receive direct feedback for knowledge enhancement. Being part of the human society is a widely studied interesting challenge~\cite{leite2013social}. Human streamers mainly attract audiences through their personality, thus attention should be given to Thyia's audience interactions. Further studies will look more into how human streamers interact with their audience, to enhance Thyia's abilities in this area (e.g. what information to present, what conversation it could be involved in etc.).

There is an important factor to take into account: the internet troll phenomenon. On the internet, many humans are generally inclined to give purposefully misleading information. When faced with an AI eager to learn from what the internet has to offer, we speculate these humans to be even more eager to fool our system. Therefore, content filtering, moderation and maintenance are necessary to ensure the system does not fall into traps. These will be further discussed in Section~\ref{sec:ethics}.





\section{Ethical implications} \label{sec:ethics}

A large limitation of the project is its possible ethical implications. The fact that Thyia would be open to outside world interaction poses a problem. The ideal scenario is the interaction taking place in a controlled environment where it would be ensured that the tasks the AI is given to solve are not ethically questionable. However, as discussed at the end of Section~\ref{subsec:interaction}, the internet is nowhere close to a controlled environment and humans interacting with the system could be supplying various malicious information:
\begin{itemize}
    \item Suggestions for unethical strategies (e.g. destroying the human race before running to the finish line).
    \item Unethical game proposals: the games sent to the system to play could contain harmful content, hate speech or unethical themes such as killing a particular race.
    \item Malicious injections taking advantage of the natural language parser to generate unexpected and harmful behaviour (e.g. teaching the agent to reply in a harmful way to the humans interacting with the system).
\end{itemize}

There have already been cases of abuse towards interactive AI. A prime example is Tay, Microsoft's chatter bot which was given open communication via Twitter, with the result being the Twitter community teaching the chatbot to become offensive and racist in only 16 hours~\cite{schlesinger2018let}. The benefit of such incidents is that subsequent attempts at general-public interaction include safety precautions against malicious intent. Moderation and content filtering are, therefore, very important to integrate within our human interaction module. One form of filtering for textual and speech-based content is sentiment analysis~\cite{jain2019systematic}, which we would use to identify possibly harmful messages received by Thyia before she gets to process them and react accordingly. However, even though research in the area of textual sentiment analysis is plentiful, it is harder to apply the same tools for game content: how could one identify if a given game is unethical? We suggest this as an interesting path for further work.


\section{Conclusions} \label{sec:end}

In this paper we present our vision for Thyia, a large system comprised of several modules with different focus and functionality. The core concept is based on the idea of an Artificial Intelligence entity which continuously plays games, using planning to solve the problems imposed and learning to improve its performance over time. The learning can be seen two-fold: in terms of the knowledge of the system and its ability to solve problems of diverse complexities and interact with dynamic environments; but also in terms of adjusting its parameters and structure so as to evolve and adapt to the worlds it encounters. We also see interaction with the real world as an aspect of key importance: an AI entity able to interact with humans in meaningful ways (such as direct communication, knowledge exchange, experience sharing) is much more interesting to study than algorithms which only exist in their constrained environments. 

We have described several limitations of the system throughout the paper and we highlight that combining many different components is bound to raise several issues. Planning methods require models of the game worlds to be able to simulate ahead and Rolling Horizon Evolutionary Algorithms, as well as Neural Networks, offer great flexibility and customisation with the risk of manually tailored parameters may not be the best choice. Hyper-parameter optimisation methods such as the N-Tuple Bandit Evolutionary Algorithm can help find good combinations of parameters, yet the game-playing and learning methods need to be able to cope with a change in parameters across the system and limit their assumptions. Lastly, human interaction brings natural language processing complications and raises some ethical concerns that should be taken into account when building a system such as Thyia.


There are many ways in which the concept of this system could be expanded even further. Knowledge representation is an interesting aspect to consider: we might want the knowledge of our system to be stored in a way such that it is more easily interpreted than a Neural Network. To this extent, Hierarchical Knowledge Bases~\cite{apeldoorn2017towards} seem like a natural addition.

Furthermore, in order for the system to be a truly general game-player, it should be able to play games even when a game model is not provided. Learning forward models in the general game-playing context is an active area of research, with several impressive advances~\cite{hafner2018planet,buesing_2018,dockhorn2018forward,lucas2019local}. With the addition of such a module, we speculate the system could even receive games from external sources (thus not adhering to any accidental assumptions included in the building of the system) and learn how to play them.

An important aspect to be considered in future work is the evaluation of the system. Given its complexities and `always-on' characteristics, evaluation would have to be done based on the system's outputs to user queries to analyse its current knowledge and skills. Given its lack of compatibility with traditional benchmarks and evaluation systems, new benchmarks could be considered for continuous evaluation of complex systems such as Thyia. Additionally, it would be interesting to explore the algorithm's ability to produce not only `intelligent' game-play, but also meaningful, creative, fun or inspiring experiences for the players it interacts with.


Lastly, we acknowledge analytics as an important possible addition. With AI systems becoming more intelligent, but also large black boxes, it is important to be able to understand their thinking process that leads to certain behaviours. Answering the question of why a decision was made would be arguably more important than making the right decision. There have been several approaches taken to extract features from a planning agent's own experience while playing a game~\cite{gaina2018win}, as well as to visualise these features to give a better insight into the agent's decisions~\cite{gaina2018vertigo}.


\section*{Acknowledgment}

This work was funded by the EPSRC CDT in Intelligent Games and Game Intelligence (IGGI) EP/L015846/1.

\bibliographystyle{IEEEtran}
\bibliography{IEEEabrv,main}

\end{document}